\documentclass[11pt]{article}

\usepackage[final]{acl}
\usepackage{times}
\usepackage{latexsym}
\usepackage{amsmath}
\usepackage{amssymb}
\usepackage{booktabs}
\usepackage{tikz}
\usetikzlibrary{arrows.meta,backgrounds,calc,fit,positioning}
\usepackage[T1]{fontenc}
\usepackage[utf8]{inputenc}

\title{Externalizing Requirement-to-Repair Artifacts as Observable Traces for LLM-Based Program Repair}

\author{
  Zewen Tao \and Shin-nosuke Ishikawa \\
  Rikkyo University \\
  \texttt{26vr047w@rikkyo.ac.jp}, \texttt{shinnosuke-ishikawa@rikkyo.ac.jp}
}

\begin{document}
\frenchspacing
\maketitle

\begin{abstract}
Repository-level repair requires not only correct patches but also inspectable records that explain how issue requirements are translated into code changes and post-edit evidence. We contribute \textsc{THEMIS}, a stage-aware repair workflow that externalizes this requirement-to-repair process through semantic interpretation, a runtime requirement–code graph, graph-derived Developer guidance, retained repair rationale and patches, and post-edit audit records. A retrospective audit of 300 SWE-bench Lite cases demonstrates that these artifacts provide broad support for cross-stage inspection: a complete Developer rationale is available for 288 cases, and 214 cases (71.3\%) retain a complete audited field set connecting the selected stages. The retained records further enable systematic measurement of cross-stage correspondence: target symbols recur in 62.6\% of Developer rationales and in 62.8\% of patches, rising to 75.8\% when related symbols are included. In a paired 100-case comparison, the relational workflow resolves 19 cases versus 9 for the direct same-input condition; because the two conditions also differ in Analyzer output, graph-derived distillation, and Judge records, we report this as preliminary, workflow-level evidence rather than a causal effect of the graph component. Together, these results show that \textsc{THEMIS} makes otherwise implicit requirement-to-repair transitions inspectable, enabling systematic examination of how repair decisions persist, align, and evolve across stages.
\end{abstract}

\section{Introduction}
\label{sec:introduction}

Automatically repairing real software issues requires more than generating plausible code edits. A GitHub issue may describe observed behavior, expected behavior, failing tests, examples, and implicit domain constraints. Benchmarks such as SWE-bench make this setting concrete by asking systems to resolve real repository-level issues from natural-language reports and code contexts \citep{jimenez-etal-2024-swebench}. LLMs have expanded the space of possible repairs, but repository-level automated program repair remains difficult because a system must jointly understand the report, localize relevant code, infer the intended behavior, and produce a minimal patch that satisfies the test oracle \citep{monperrus-2018-automatic-repair-bibliography,jiang-etal-2023-code-lms-apr}.

The central information problem arises between repair stages: a system must carry issue requirements into code edits and post-edit audit while preserving records that distinguish each stage and make their relations inspectable. The design therefore begins with semantic interpretation of the issue, makes the resulting requirement-code relations explicit, distills selected relations into a bounded rationale and edit, and retains post-edit evidence for audit. This why-chain connects the system's stages: semantic interpretation supplies the meaning needed to relate requirements to code; explicit relation state makes those connections inspectable; bounded rationale and editing preserve a focused execution record; and post-edit evidence shows what can be examined after the change. This paper retrospectively and descriptively audits retained records from 300 SWE-bench Lite cases. Conversational repair systems add feedback from tests or prior attempts \citep{xia-zhang-2023-conversational-apr,xia-etal-2024-chatrepair}; this paper examines one repair pass.

We ask: \textbf{RQ1: What stage-specific information can be externalized to make the transition from issue requirements to code edits and post-edit audit inspectable, and where do continuity gaps arise?} \textbf{RQ2: What observable correspondences and disagreements arise among target localization, repair rationale, generated edits, and post-edit evidence in repository-level repair?} \textbf{RQ3: What preliminary outcome differences are observed between a relational repair workflow and a direct same-input repair condition?} RQ1 is operationalized through selected-field availability and continuity gaps; RQ2 through retained post-edit target summaries, Developer rationale, and patch-text lexical comparisons; and RQ3 through a paired 100-case workflow comparison.

We present \textsc{THEMIS} as one instantiation of an explicit selected stage-aware audit projection.\footnote{\raggedright Code: \url{https://github.com/ForeverMJ/THEMIS}} It constructs repair context from a requirement-code graph, then retains selected graph-derived summaries, Developer rationale, patches, and post-edit diagnostics. Its Developer receives a textual distillation. The projection is \emph{inspectable} through its readable retained fields and \emph{traceable} through the workflow's asserted links among them; the study audits those selected serialized records in an augmented-input, single-revision configuration.

This paper makes three contributions.
\begin{enumerate}
    \item \textbf{Stage information and continuity for RQ1:} We define the selected stage-aware projection and report retained-field availability and continuity gaps in Table~\ref{tab:trace-availability}.
    \item \textbf{Observable correspondence and disagreement for RQ2:} We characterize lexical correspondences and disagreements among retained post-edit target summaries, Developer rationale, and patches in Table~\ref{tab:lexical-characterization} and two illustrative cases in Section~\ref{subsec:trace-cases}.
    \item \textbf{Preliminary workflow outcomes for RQ3:} We compare 100 paired cases between the relational workflow and a direct same-input condition: 19/100 versus 9/100 official resolutions and 89/100 versus 43/100 non-empty patches. This workflow-level comparison spans conditions with distinct Analyzer output, graph-derived distillation, and Judge records.
\end{enumerate}

\section{Related Work}
\label{sec:related-work}

\paragraph{Automated program repair.}
Automated program repair is commonly formulated as generate-and-validate search: candidate edits are generated and checked against tests, specifications, crashes, or semantic constraints \citep{monperrus-2018-automatic-repair-bibliography}. Classical systems include search-based repair such as GenProg \citep{weimer-etal-2009-genprog}, empirical studies of test-suite repair \citep{legoues-etal-2012-systematic-apr}, learned patch priors such as Prophet \citep{long-rinard-2016-prophet}, and semantic/symbolic systems such as SemFix, Angelix, and Nopol \citep{nguyen-etal-2013-semfix,mechtaev-etal-2016-angelix,xuan-etal-2017-nopol}. THEMIS follows the constraint-guided repair tradition, but its constraints are derived from issue text and represented as requirement-code graph relations.

\paragraph{LLM-based repository repair.}
LLMs have shifted repair from hand-designed templates toward flexible code generation. Prior work studies LLMs for bug fixing and repository-level issue resolution, including conversational repair, SWE-bench agents, and agentless repair pipelines \citep{sobania-etal-2023-chatgpt-bugfixing,xia-zhang-2023-conversational-apr,xia-etal-2024-chatrepair,jimenez-etal-2024-swebench,yang-etal-2024-swe-agent,xia-etal-2024-agentless}. THEMIS complements these systems by structuring the information given to a Developer model through explicit issue requirements and code mappings.

\paragraph{Graphs and semantic guidance for code.}
Graph representations have long been used to encode code structure beyond token sequences. Code property graphs combine syntax, control flow, and dependence information for program analysis \citep{yamaguchi-etal-2014-code-property-graphs}; neural code models use syntactic and semantic graph relations for code understanding \citep{allamanis-etal-2018-program-graphs,alon-etal-2019-code2seq,alon-etal-2019-code2vec,guo-etal-2021-graphcodebert}. THEMIS adds issue-derived requirement nodes and violation/advisory edges, using the graph as both a code representation and a repair-context state. The selected projection declares stage roles and selected relations explicitly, and structured logging can make the same declarations. The direct same-input baseline compares the reported workflow conditions across their representation formats.

\paragraph{Software traceability and agent observability.}
A separate line of work studies traceability directly: software and systems traceability surveys and IR-based trace-link recovery connect requirements to downstream artifacts and evaluate how reliably such links can be recovered \citep{cleland-huang-etal-2012-traceability,borg-etal-2014-traceability-mapping}. Recent surveys of LLM agents similarly call for unified schemas that record how evidence, tool calls, and intermediate reasoning shape an agent's final action, so that execution provenance can be audited after the fact \citep{wang-etal-2026-agent-traces-survey}. THEMIS targets this same auditability goal for repository-level repair specifically, externalizing a requirement-to-repair trace rather than a general-purpose agent execution log.

\section{System Architecture}
\label{sec:system-architecture}

\begin{figure*}[t]
\centering
\resizebox{0.99\textwidth}{!}{%
\begin{tikzpicture}[
    font=\sffamily\scriptsize,
    flow/.style={-{Stealth[length=2.2mm]}, line width=0.55pt},
    trace/.style={-{Stealth[length=2.2mm]}, line width=0.55pt, dashed},
    box/.style={draw=black!75, rounded corners=1.5pt, align=center,
        inner sep=4pt, minimum height=9mm, fill=white},
    runtime/.style={box, dashed},
    persisted/.style={box, fill=black!12},
    stage/.style={draw=black!30, rounded corners=2pt, fill=black!3,
        inner sep=5pt},
    stage title/.style={font=\sffamily\bfseries\scriptsize, fill=white,
        inner xsep=3pt, inner ysep=1pt}
]
\node[box, text width=20mm] (input) {Issue text\\+ selected files};
\node[box, text width=27mm, right=7mm of input, yshift=10mm] (analyzer)
    {\textbf{Advanced Code Analyzer}\\findings, recommendations};
\node[runtime, text width=27mm, right=7mm of input, yshift=-10mm] (basegraph)
    {\textbf{Enhanced Graph Manager}\\baseline graph: requirements, code, mappings, violations};
\node[box, text width=30mm, right=8mm of analyzer, yshift=-7mm] (distill)
    {\textbf{Developer-facing distillation}\\priorities, targets, plans, ingredients};

\node[box, text width=22mm, right=12mm of distill] (developer)
    {\textbf{Developer}\\single bounded edit};
\node[persisted, text width=25mm, below=12mm of developer] (devrecord)
    {Developer rationale\\+ patch record};

\node[runtime, text width=23mm, right=15mm of developer, yshift=7mm] (rebuild)
    {\textbf{Rebuilt graph}\\post-edit state};
\node[box, text width=21mm, right=7mm of rebuild] (judge)
    {\textbf{Judge}\\conflict checks};
\node[persisted, text width=28mm, below=12mm of judge] (postrecord)
    {Conflict history\\+ repair brief};
\node[persisted, text width=29mm, below=12mm of distill] (summary)
    {Selected graph-derived\\summaries};

\draw[flow] (input) -- (analyzer);
\draw[flow] (input) -- (basegraph);
\draw[flow] (analyzer) -- (distill);
\draw[flow] (basegraph) -- (distill);
\draw[flow] (distill) -- (developer);
\draw[flow] (developer) -- node[above, font=\sffamily\tiny] {patch} (rebuild);
\draw[flow] (rebuild) -- (judge);
\draw[flow] (distill) -- (summary);
\draw[flow] (developer) -- (devrecord);
\draw[flow] (judge) -- (postrecord);
\draw[trace] (summary) -- (devrecord);
\draw[trace] (devrecord) -- (postrecord);

\coordinate (boundarytop) at ($(developer.east)!0.5!(rebuild.west)+(0,17mm)$);
\coordinate (boundarybottom) at ($(developer.east)!0.5!(rebuild.west)+(0,-25mm)$);
\draw[densely dashdotted, black!80] (boundarytop) -- (boundarybottom);
\node[anchor=south east, xshift=-1mm, font=\sffamily\scriptsize, align=right]
    at (boundarytop) {single-revision\\boundary};

\begin{scope}[on background layer]
\node[stage, fit=(input)(analyzer)(basegraph)(distill)(summary)] (prestage) {};
\node[stage, fit=(developer)(devrecord)] (repairstage) {};
\node[stage, fit=(rebuild)(judge)(postrecord)] (poststage) {};
\end{scope}
\node[stage title, anchor=south west] at (prestage.north west) {PRE-EDIT GUIDANCE};
\node[stage title, anchor=south west] at (repairstage.north west) {REPAIR EXECUTION};
\node[stage title, anchor=south west] at (poststage.north west) {POST-EDIT AUDIT};

\node[below=7mm of devrecord, text width=160mm, align=center,
    font=\sffamily\scriptsize] (legend)
    {\textbf{Legend:} solid/dashed border = component or runtime-only graph; gray fill = persisted artifact;
    solid/dashed arrow = reported flow or asserted trace link.};
\end{tikzpicture}%
}
\caption{THEMIS artifact flow in the reported single-revision configuration. The Advanced Code Analyzer supplies soft semantic findings; the Enhanced Graph Manager makes requirements, targets, and constraints explicit; the Developer consumes readable distilled context and records its rationale and patch; and the Judge records post-edit conflict history and repair briefs. Dashed-border boxes are runtime-only graph states; shaded boxes are selected persisted artifacts. Solid arrows show reported data flow. Persisted artifacts and dashed links denote recorded, workflow-asserted relations. Judge outputs follow the edit and serve exclusively as post-edit audit records in the single-revision workflow.}
\label{fig:themis-architecture}
\end{figure*}
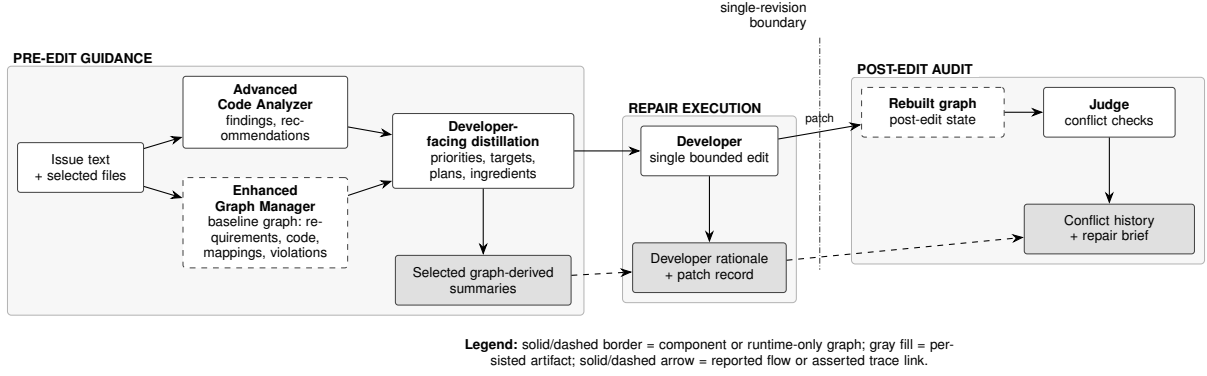

THEMIS contains four modules: an Advanced Code Analyzer, an Enhanced Graph Manager, a Developer, and a Judge. Its graph implements explicit requirement-code relations within the reported workflow. Together the modules define an explicit, selected stage-aware audit projection that captures pre-edit interpretation and guidance, repair execution and rationale, and post-edit evidence. This separation serves the method's inspectability objective: the analyzer keeps issue-level semantic hypotheses distinct from structural relation claims; the graph manager exposes selected requirement-target-constraint relations as inspectable anchors; the Developer retains the guidance-to-rationale-to-patch transition as an inspectable execution record; and the rebuilt graph and Judge retain post-edit evidence as a distinct audit stage. Figure~\ref{fig:themis-architecture} preserves this ordering across the reported single-revision boundary: pre-edit guidance precedes one repair execution, followed by post-edit audit.

\subsection{Advanced Code Analyzer}
\label{subsec:advanced-analyzer}

The Advanced Code Analyzer is introduced to retain issue-level semantic hypotheses from the issue text and local code separately from structural relation claims. This separation keeps the two forms of pre-edit information inspectable and the semantic findings distinct from graph edges. It receives the issue text, selected target files, optional code context, and analysis options. It performs bug classification, semantic extraction, context enhancement, concept mapping, pattern matching, and multi-round LLM reasoning. Its outputs are findings, recommendations, confidence scores, strategy metadata, and usage information when available. These outputs enter the Developer prompt as \emph{semantic priors}: they suggest likely fault mechanisms or repair directions, remain distinct from structural targets, and occupy a separate pre-edit record category.

\subsection{Enhanced Graph Manager}
\label{subsec:graph-manager}

The Enhanced Graph Manager is introduced to expose selected requirement-target-constraint relations as inspectable anchors for pre-edit distillation. Its expected benefit is that these selected relations and code neighborhoods can be examined as explicit structural records. The Enhanced Graph Manager builds the requirement-code graph. A structural extractor parses Python source into code nodes such as functions, classes, and variables. A semantic injector decomposes issue text into \texttt{RequirementNode}s and maps them to candidate code nodes. A dependency tracer adds code-neighborhood information. A violation flagger analyzes requirement-code pairs and produces \texttt{ViolationEdge}s.

The graph distinguishes blocking and non-blocking guidance. High-signal blocking mismatches are recorded as \texttt{VIOLATES} edges. Lower-confidence or non-blocking signals are recorded as \texttt{ADVISORY} edges. The graph manager then summarizes these edges in an analysis report containing violation counts, prioritized violations, evidence scores, confidence values, and dependency statistics.

\subsection{Developer}
\label{subsec:developer}

The Developer consumes selected actionable context to produce one bounded edit. It is introduced to preserve the guidance-to-rationale-to-patch transition as an inspectable execution record. Textual distillation is a pre-edit artifact that presents selected priorities, targets, plans, and local ingredients in readable form. The Developer rationale and patch are execution artifacts that retain the transition from guidance to execution. The Developer receives the current file map, original requirements, Advanced Analyzer findings, graph-derived violation priorities, target symbols, repair operator plans, and local code ingredients. Graph artifacts reach the Developer as concise textual guidance. The Developer produces bounded edits in either search/replace form or symbol-rewrite form. The runner accepts edits that satisfy file, syntax, and substantive-change checks.

\subsection{Judge}
\label{subsec:judge}

The Judge creates a post-edit record from the graph rebuilt after the single edit, separating remaining blocking conflicts from advisory findings. It is introduced to retain post-edit evidence as a distinct audit stage in the single-revision workflow. This timing positions the post-edit conditions and repair brief for inspection after Developer execution. The Judge consumes a revised graph, the requirements, and optionally the baseline graph. Its hard check scans graph edges of type \texttt{VIOLATES} and \texttt{ADVISORY}, separates blocking conflicts from advisory findings, and builds a conflict report. It also constructs a repair brief with a target symbol, related symbols, issue summary, expected behavior, and minimal-change hint. In the current experiments, these outputs serve exclusively as post-edit audit records after the workflow's one Developer revision.

\subsection{First-Pass Repair Context}
\label{subsec:first-pass-context}

The first-pass context implements the pre-edit portion of the design chain: it keeps raw requirements and files, soft semantic clues, and explicit graph-derived constraints distinguishable while presenting their selected forms together for a bounded edit. The first Developer pass receives three kinds of information. First, raw inputs provide the requirements and selected source files. Second, the Advanced Code Analyzer provides semantic findings and recommendations. Third, the graph manager provides baseline graph-derived constraints: prioritized requirement-code violations, target symbols, repair operator plans, and local code ingredients. Judge conflict reports and repair briefs are generated after a Developer edit and graph rebuild, where they serve as post-edit audit records.

\subsection{Inspectable Trace Artifacts}
\label{subsec:trace-artifacts}

The architecture serializes selected artifacts so that pre-edit interpretation and targets, repair execution, and post-edit audit remain distinct for inspection. Canonical logs retain selected requirement and target summaries, Developer rationale, patch, and post-edit diagnostics. During analysis and graph rebuild, the runtime graph holds requirement nodes, code nodes, mappings, and conflict relations; the selected serialization defines the audit scope. The Developer receives textual distillation of selected graph-derived items. The audit measures availability, surface overlap, and quality flags in the retained records. Table~\ref{tab:artifact-timing} distinguishes their timing, consumers, and persistence.

\begin{table*}[t]
\centering
\small
\begin{tabular}{p{0.21\linewidth}p{0.19\linewidth}p{0.19\linewidth}p{0.30\linewidth}}
\toprule
Artifact & Producer and timing & Consumer & Persistence in canonical logs \\
\midrule
Runtime requirement-code graph & Graph Manager before edit; rebuilt after edit & Distillation and Judge & Runtime graph during analysis; selected summaries are serialized \\
Developer-facing distillation & Graph Manager before the single Developer edit & Developer & Selected textual guidance derived from graph artifacts \\
Developer rationale and patch & Developer during the single edit & Runner and retrospective audit & Rationale fields and patch record when available \\
Repair brief and conflict history & Judge after edit and graph rebuild & Retrospective audit & Selected post-edit summaries and conflict history \\
Official outcome & SWE-bench harness after submission & Retrospective audit & Harness resolution record \\
\bottomrule
\end{tabular}
\caption{Artifact timing, consumers, and persistence. The repair brief and Judge diagnostics are post-edit records in the reported single-revision run.}
\label{tab:artifact-timing}
\end{table*}

We define an observable trace schema as $\tau=(r,t,d,p,q)$, where $r$ is a serialized requirement summary, $t$ is a selected target summary, $d$ is Developer rationale, $p$ is the patch record, and $q$ is a post-edit diagnostic. A trace is \emph{inspectable} when these stored artifacts are readable. It is \emph{traceable} when the workflow asserts links among them. This tuple captures selected workflow-asserted records and relations for availability, surface-overlap, and quality-flag analysis.

\section{Experimental Setup}
\label{sec:experimental-setup}

We evaluate THEMIS on SWE-bench Lite, a 300-instance subset of real GitHub issue-resolution tasks from Python repositories \citep{jimenez-etal-2024-swebench}. Each instance provides a base commit, issue text, optional hints, and tests used by the official evaluation harness. THEMIS consumes the issue text and selected source files, produces a patch as a git diff, and writes predictions in the JSONL format expected by the harness.

All reported runs use gpt-5.1-codex-mini for the repair workflow and the Advanced Code Analyzer, chosen for its balance of coding capability and inference cost relative to larger code-specialized models; we leave comparison across model families and scales to future work. The main run uses the integrated configuration with Advanced Analysis, graph construction, one Developer revision, graph rebuild, and post-edit Judge audit. Repair briefs and Judge diagnostics are generated after that edit and rebuild. Thus, the evaluation covers a single graph-guided repair pass and its post-edit audit records.

The reported condition is explicitly augmented. THEMIS uses \texttt{FAIL\_TO\_PASS} test identifiers during file selection and requirement construction. The runner extracts explicit Python paths from issue text and failing-test metadata when available, otherwise searches the repository for issue terms and ranks candidate files. The requirement string concatenates the problem statement, available hints, \texttt{FAIL\_TO\_PASS} identifiers, and any manually supplied semantic contracts. The reported evidence applies to this augmented-input SWE-bench condition; issue-only systems are a future comparison target. Selected presets add bounded context, retrieval, or semantic-contract information for controlled slices.

We operationalize the three general research questions from retained records. RQ1 measures selected-field retention and gaps in the retained summary chain. RQ2 measures string-level surface correspondence or disagreement among the retained post-edit target summary, the last Developer rationale, and patch text. RQ3 is a paired 100-case direct same-input workflow comparison. All 100 pairs match instance IDs, gpt-5.1-codex-mini, seed 42, one Developer revision, the requirement-string builder, the selected-file strategy, and the actual selected files. Advanced Analyzer output, graph-derived distillation and relations, and Judge records occur exclusively in the relational workflow; the two conditions use different token budgets and interfaces. Full-run lineage predates \texttt{input\_\allowbreak protocol} and is established retrospectively from experiment metadata and baseline declarations. This comparison supports preliminary workflow-level interpretation across the reported condition differences. The target lineage is exactly post-edit \texttt{meta.\allowbreak repair\_\allowbreak brief.\allowbreak target\_\allowbreak symbol}: generated after the edit, it serves as a retrospective lexical target, distinct from pre-edit guidance and patch generation. The rationale concatenates the last recorded \texttt{hypothesis\_\allowbreak root\_\allowbreak cause}, \texttt{expected\_\allowbreak invariant}, and \texttt{patch\_\allowbreak strategy}; JSON \texttt{null} marks missing values, and \texttt{None} is a string value. Canonical patch text is \texttt{model\_\allowbreak patch}.

For matching, we apply Unicode-aware \texttt{casefold()}, replace \texttt{::} with \texttt{.}, and match either the qualified or leaf form as a substring of casefolded rationale or \texttt{model\_\allowbreak patch}. An expanded patch row is a hit when \texttt{target\_\allowbreak symbol} or any non-empty \texttt{related\_\allowbreak symbols[]} entry matches \texttt{model\_\allowbreak patch} under that rule. This string-level rule measures surface correspondence among the retained records. The rationale cohort of 214 requires repair-brief core, non-empty \texttt{conflict\_\allowbreak metrics\_\allowbreak history}, and non-empty last Developer rationale fields. The patch cohort of 207 requires repair-brief core and non-empty canonical \texttt{model\_\allowbreak patch}; the 205-case intersection is the joint cohort. Low confidence is \texttt{meta.\allowbreak repair\_\allowbreak brief.\allowbreak confidence} $\leq 0.2$; a generic issue summary is trimmed exact membership in \texttt{No clear relationship found} or \texttt{No validation functions found in codebase}. Both flags use the 222 repair-brief-core cases as their denominator. We report official-harness resolved rate, resolved rate among non-empty-patch instances, and patch generation rate. Patch production records an edit, while official resolution records a benchmark-passing edit.

\section{Results}
\label{sec:results}

\subsection{Availability of Stage-Specific Repair Records}
\label{subsec:trace-availability}

RQ1 asks whether selected records form a complete field chain available for inspection. We check complete chains because an auditor needs all three retained field groups to read the cross-stage record together. We therefore audited exactly nine existing default-run log directories covering the 300 cases. Conflict history is a non-empty \texttt{conflict\_\allowbreak metrics\_\allowbreak history} list. Developer rationale requires non-empty \texttt{hypothesis\_\allowbreak root\_\allowbreak cause}, \texttt{expected\_\allowbreak invariant}, and \texttt{patch\_\allowbreak strategy}. Repair-brief core requires non-empty \texttt{requirement\_\allowbreak id}, \texttt{target\_\allowbreak symbol}, and \texttt{expected\_\allowbreak behavior}. An audited field set is complete when all three field groups are retained.

\begin{table}[t]
\centering
\small
\begin{tabular}{lr}
\toprule
Serialized artifact & Available cases \\
\midrule
Conflict history & 300/300 (100.0\%) \\
Complete Developer rationale & 288/300 (96.0\%) \\
Repair-brief core & 222/300 (74.0\%) \\
Complete audited field set & 214/300 (71.3\%) \\
\bottomrule
\end{tabular}
\caption{Retrospective availability of selected serialized artifacts across exactly nine default-run log directories.}
\label{tab:trace-availability}
\end{table}

For RQ1, Table~\ref{tab:trace-availability} reports conflict history for all 300 cases and complete field-chain coverage for 214/300 (71.3\%). The 86 remaining cases identify the retained fields that require attention for joint cross-stage inspection. The 214/300 figure reports selected-field coverage in the retained record set.

\subsection{Surface Correspondence and Record-Quality Flags}
\label{subsec:lexical-characterization}

For RQ2, Table~\ref{tab:lexical-characterization} measures lexical overlap to make visible target recurrence in rationales and patches under the stated string rule. It reports descriptive quality flags separately to make uncertainty and limited specificity in the retained summaries visible. Under the stated string rule, recorded targets appear in 134/214 rationales and 130/207 patches; allowing related symbols raises patch overlap to 157/207. These counts characterize lexical surface correspondence and disagreement among retained records.

\begin{table}[t]
\centering
\small
\begin{tabular}{lr}
\toprule
Characterization & Cases \\
\midrule
Target in rationale & 134/214 (62.6\%) \\
Target in patch & 130/207 (62.8\%) \\
Target or related symbol in patch & 157/207 (75.8\%) \\
Low-confidence repair brief & 131/222 (59.0\%) \\
Generic issue summary & 138/222 (62.2\%) \\
\bottomrule
\end{tabular}
\caption{Lexical alignment and quality flags for selected serialized artifacts.}
\label{tab:lexical-characterization}
\end{table}

Within the 205-case joint cohort, both rationale and patch match in 110 cases; the remaining 95/205 distribute across 18 rationale-only matches, 19 patch-only matches, and 58 records in the zero-overlap cell. Their respective resolved counts are 29 (26.4\%), 8 (44.4\%), 2 (10.5\%), and 8 (13.8\%). The matrix makes visible both string-surface agreement and disagreement. Across the repair-brief-core cohort, 131/222 repair briefs are low confidence and 138/222 have generic issue summaries. Together, the overlap and flag results make correspondence, mismatch, uncertainty, and limited specificity visible in the retained records.

\subsection{Two Illustrative Stage-Specific Record Cases}
\label{subsec:trace-cases}

The two cases make RQ2 patterns inspectable beyond aggregate counts. They serve as illustrative records that connect the aggregate surface measures to one readable correspondence and one record-level mismatch.

The resolved case \texttt{django\_\_\allowbreak{}django-11620} is the correspondence case. It selected \texttt{django/views/debug.py}; its repair brief identifies \texttt{REQ-029}, targets \texttt{resolve}, and lists the related symbol \texttt{technical\_\allowbreak404\_\allowbreak response}; the Developer rationale states that the relevant path should catch \texttt{Http404} alongside \texttt{Resolver404}; and the patch makes that change. The official harness outcome is resolved. The retained records expose readable surface correspondence.

The unresolved case \texttt{django\_\_\allowbreak{}django-16139} is the disagreement case. It selected \texttt{django/contrib/auth/forms.py}; the Developer rationale targets \texttt{UserChangeForm} and a \texttt{\_to\_\allowbreak{}field}-aware URL; in contrast, the graph repair brief targets \texttt{SetPasswordForm}, reports confidence 0.2, and states ``No clear relationship found.'' The patch modifies \texttt{UserChangeForm}, and the official harness outcome is unresolved. The retained records expose an inspectable record-level target mismatch.

\subsection{Patch Production and Official Outcomes}
\label{subsec:overall-results}

This subsection reports outcome context through inspectable patch production and benchmark success, which answer different questions. In the 300-instance augmented-input, single-revision run, 279 cases produce non-empty patches (93.0\%), 58 of 300 are officially resolved (19.3\%), and 58/279 non-empty-patch cases are resolved (20.8\%). The distinction keeps produced patch records and benchmark-passing outcomes separately visible. These results provide descriptive outcome context for the reported condition and identify causal-effect validation as a future target.

\subsection{Descriptive Outcome Context by Recorded Condition}
\label{subsec:outcome-associations}

We compare recorded-condition strata for descriptive official-outcome context; denominators require their respective available fields.

\begin{table}[t]
\centering
\small
\begin{tabular}{lr}
\toprule
Condition & Resolved \\
\midrule
Complete audited field set & 47/214 (22.0\%) \\
Incomplete audited field set & 11/86 (12.8\%) \\
Target in rationale & 37/134 (27.6\%) \\
Target absent from rationale & 10/80 (12.5\%) \\
Target in patch & 31/130 (23.8\%) \\
Target absent from patch & 17/77 (22.1\%) \\
\bottomrule
\end{tabular}
\caption{Descriptive official outcome associations for available artifact subsets.}
\label{tab:outcome-associations}
\end{table}

The separations are 22.0\% versus 12.8\% for complete versus incomplete field chains, 27.6\% versus 12.5\% for rationale-target presence versus absence, and 23.8\% versus 22.1\% for patch-target presence versus absence. Rationale-target presence separates these descriptive strata more than patch-target presence. These associations are correlational and likely confounded by case difficulty: cases with incomplete field chains may simply be harder cases where the pipeline broke down for reasons that also make them harder to resolve.

\subsection{Paired Same-Input Workflow Comparison}
\label{subsec:same-input-rq3}

RQ3 compares the paired 100-case direct same-input workflow conditions in Table~\ref{tab:same-input-rq3} to place their outcomes in workflow-level context. THEMIS resolves 19/100 versus 9/100; the paired cells are 9 both resolved, 10 THEMIS-only resolved, 0 direct-only resolved, and 81 unresolved in both conditions. The paired risk difference is +0.10, with an approximate 95\% confidence interval of [0.0412, 0.1588]. The exact two-sided McNemar test uses the 10 discordant pairs, 10 THEMIS-only and 0 direct-only, and yields $p = 0.002$. Non-empty patches occur in 89/100 THEMIS cases and 43/100 direct-condition cases. These results provide preliminary workflow-level evidence of a difference between the two conditions.

\begin{table}[t]
\centering
\small
\begin{tabular}{lr}
\toprule
Paired outcome or patch aggregate & Cases \\
\midrule
THEMIS resolved & 19/100 \\
Direct condition resolved & 9/100 \\
Both resolved & 9 \\
THEMIS-only resolved & 10 \\
Direct-only resolved & 0 \\
Neither resolved & 81 \\
THEMIS non-empty patch & 89/100 \\
Direct condition non-empty patch & 43/100 \\
\bottomrule
\end{tabular}
\caption{Paired 100-case official outcomes and non-empty patch aggregates for the same-input workflow comparison. The paired risk difference is +0.10, the approximate 95\% confidence interval is [0.0412, 0.1588], and the exact two-sided McNemar $p$ is 0.002.}
\label{tab:same-input-rq3}
\end{table}

This is preliminary workflow-level evidence across conditions with distinct Analyzer output, graph-derived distillation and relations, Judge records, interfaces, and token budget. The observed difference supports workflow-level interpretation.

\section{Discussion}
\label{sec:discussion}

\paragraph{RQ1: externalized stage information exposes continuity gaps.}
The selected projection externalizes conflict history for 300/300 cases, complete Developer rationale for 288/300, and repair-brief core fields for 222/300. Evaluating their joint availability matters because stage separation supports inspection when the retained pre-edit, edit, and post-edit records can be read together. All three field groups are jointly available in 214/300 cases; the remaining 86 cases locate continuity gaps for targeted serialization improvements. Thus, the availability evaluation identifies the retained records that sustain the stage-aware audit design and the fields that require attention within the selected serialization.

\paragraph{RQ2: separately retained records make lexical correspondence and disagreement observable.}
Separately retaining targets, rationales, and patches makes the lexical comparison informative because it exposes agreement and disagreement among records as distinct observations. Recorded targets occur in 134/214 Developer rationales and in 130/207 patches under the stated string rule; in the 205-case joint cohort, 95/205 cases divide among rationale-only, patch-only, and zero-overlap cells. The stage-specific records therefore make surface correspondence, disagreement, and target mismatches inspectable. The quality flags complement this comparison as design signals: among the 222 repair-brief cores, 131/222 are low confidence and 138/222 have generic issue summaries, making uncertainty and limited specificity visible for review. They remain record-surface observations under the stated lexical proxy, separate from RQ3 workflow outcomes, and quality-improvement validation remains a future target.

\paragraph{RQ3: the paired direct same-input comparison shows a bounded workflow-level difference.}
Comparing paired same-input cases matters because it places the two workflow conditions in a common outcome context while retaining the workflow-level interpretation. Across the paired 100 cases, \textsc{THEMIS} resolves 19/100 cases and the direct condition resolves 9/100. The discordant outcomes are 10 \textsc{THEMIS}-only resolutions and 0 direct-only resolutions. The paired risk difference is +0.10, with an approximate 95\% confidence interval of [0.0412, 0.1588], and the exact two-sided McNemar test gives $p = 0.002$. This comparison is therefore informative as preliminary evidence about the reported workflow conditions, within the confounds documented in the experimental setup and limitations.

\section{Conclusion}
\label{sec:conclusion}

Repository-level repair should be evaluated not only by whether a patch passes its tests, but also by whether the process connecting issue requirements, repair decisions, code edits, and post-edit evidence can be inspected. This paper contributes THEMIS, a stage-aware repair workflow that externalizes this process through semantic interpretation, explicit requirement–code relations, graph-derived Developer guidance, retained rationale and patch records, and post-edit diagnostics.  
The empirical results demonstrate the value of this approach in three ways. First, across 300 SWE-bench Lite cases, complete Developer rationales are retained for 288 cases, while 214 cases (71.3\%) provide a complete field chain across the selected stages. These results show that THEMIS enables joint inspection of the requirement-to-repair process at benchmark scale. Second, the retained artifacts make cross-stage correspondence systematically measurable: target symbols recur in 62.6\% of Developer rationales and 62.8\% of patches, increasing to 75.8\% when related symbols are considered. This makes it possible to observe how repair targets and decisions persist or evolve across stages rather than remaining hidden behind the final patch. Third, in the paired 100-case comparison, THEMIS resolves 19 cases compared with 9 for the direct same-input condition, providing preliminary, confounded workflow-level evidence for the relational approach; because the two conditions also differ in Analyzer output, distillation, and audit records, this comparison does not yet isolate the graph component's causal contribution.  
Taken together, these contributions establish externalized repair artifacts as a practical foundation for inspecting and evaluating the process by which repository-level repairs are produced. By turning otherwise implicit requirement-to-repair transitions into observable traces, THEMIS extends repair evaluation beyond final outcomes and enables systematic analysis of how requirements are interpreted, translated into edits, and reflected in post-edit evidence.

\section{Limitations}
\label{sec:limitations}

\paragraph{Workflow and comparison boundaries.}
The documented experiments use one Developer revision, with the Judge acting only as a post-edit audit record. RQ3 compares paired workflows that differ in Analyzer output, graph-based distillation and relations, Judge records, token budget, and interface. Lineage for this comparison is reconstructed from experiment metadata and baseline declarations. The study reports workflow-level outcomes for these configurations; future work can test representation, module, artifact, lexical-pattern, causal, and general-superiority questions.

\paragraph{Record and proxy boundaries.}
The selected serialization captures requirement and target summaries, Developer rationale, patch records, and post-edit diagnostics; the full runtime graph, complete requirements, and complete reasoning remain runtime material. Requirement decomposition, requirement-code mapping, and violation flagging are heuristic. Lexical matching measures string-level surface correspondence. Semantic correctness, trace correctness, and target-to-patch causal validation remain future targets, as does validation of the contents and relations represented by field availability and asserted links. A principled limitation applies to this proxy: the retrospective target used for RQ2 (\texttt{repair\_brief.target\_symbol}) is generated by the same Judge component after the edit and graph rebuild, rather than fixed before editing or drawn from an independent reference (e.g., the gold patch's changed lines). High lexical overlap may therefore partly reflect the Judge's agreement with the Developer's own edit rather than independently established cross-stage traceability.

\paragraph{Utility, scope, and model dependence.}
Human-utility evaluation of whether these records help readers diagnose failures, debug systems, or make better decisions remains a future validation target. Evaluation covers Python repositories in SWE-bench Lite and one model configuration, \texttt{gpt-5.1-codex-mini}. Future validation across languages, benchmarks, models, model scales, and repair settings can assess language-specific extraction and edit logic alongside workflow outcomes.

\section*{Acknowledgments}
This work was supported by JSPS KAKENHI Grant Number 24K15077.

Following the ACL Policy on AI Writing Assistance, we disclose our use of AI assistants in preparing this work: OpenCode assisted with implementing the THEMIS codebase, and Codex assisted with translating and polishing portions of the manuscript text. This writing-assistance use of Codex is unrelated to the \texttt{gpt-5.1-codex-mini} configuration evaluated as a Developer model in Section~\ref{sec:experimental-setup}. All AI-assisted content was reviewed and verified by the authors, who take full responsibility for it.

\bibliography{references}

\end{document}